\documentclass[letterpaper, 10 pt, journal, twoside]{ieeetran}  

\IEEEoverridecommandlockouts                              
\makeatletter
\def\bstctlcite{\@ifnextchar[{\@bstctlcite}{\@bstctlcite[@auxout]}}
\def\@bstctlcite[#1]#2{\@bsphack
  \@for\@citeb:=#2\do{%
    \edef\@citeb{\expandafter\@firstofone\@citeb}%
    \if@filesw\immediate\write\csname #1\endcsname{\string\citation{\@citeb}}\fi}%
  \@esphack}
\makeatother

\usepackage[vlined, ruled, linesnumbered, commentsnumbered]{algorithm2e}
\usepackage{amsmath}
\usepackage{cite}
\usepackage{color}
\usepackage{xcolor}
\definecolor{chred}{rgb}{0.8,0,0}
\definecolor{chgrey}{rgb}{0.5,0.5,0.5}
\usepackage{graphicx}
\usepackage{subcaption}
\usepackage{sidecap}
\usepackage{wrapfig}
\usepackage{dblfloatfix}
\usepackage{eqlist}
\usepackage{txfonts}
\usepackage{url}
\usepackage{footmisc}
\usepackage{xcolor}
\usepackage{booktabs}
\usepackage{balance}\usepackage{flushend}

\usepackage{multirow}
\usepackage[para,online,flushleft]{threeparttable}
\usepackage[final]{pdfpages}

\begin{document}
\title{\LARGE \bf
Tethered Tool Manipulation Planning with Cable Maneuvering 
}

\author{Daniel  S\'anchez$^{1}$, Weiwei Wan$^{1,2*}$ and Kensuke Harada$^{1,2}$
\thanks{$^{1}$Graduate School of Engineering Science, Osaka University, Japan.}%
\thanks{$^{2}$National Inst. of AIST, Japan.}%
\thanks{Contact: Weiwei Wan, {\tt\small wan@hlab.sys.es.osaka-u.ac.jp}}%
}

\maketitle

\begin{abstract}

In this paper, we present a planner for manipulating tethered tools using dual-armed robots. 
The planner generates robot motion sequences to maneuver a tool and its cable while avoiding robot-cable entanglements. 
Firstly, the planner generates an Object Manipulation Motion Sequence (OMMS) to handle the tool and place it in desired poses. 
Secondly, the planner examines the tool movement associated with the OMMS and computes candidate positions for a cable slider, 
to maneuver the tool cable and avoid collisions. Finally, the planner determines the optimal slider positions to avoid 
entanglements and generates a Cable Manipulation Motion Sequence (CMMS) to place the slider in these positions. 
The robot executes both the OMMS and CMMS to handle the tool and its cable to avoid entanglements and excess cable bending. 
Simulations and real-world experiments help validate the proposed method.

\end{abstract}

\section{Introduction}

\IEEEPARstart{T}{he} introduction of robots to manufacturing industries aims to reduce human workload, increase productivity, 
and decrease operation costs. Towards realizing these goals, robots must be able to adapt to industrial environments, 
work alongside humans, and manipulate tools to complete given tasks. Particularly, motion planning for handling tools 
represents a unique challenge for planners: The tool acts as a dynamic obstacle when it is being manipulated by the 
robot and its position and orientation during the manipulation task must be accurately computed to avoid robot-object 
and object-environment collisions. The tool manipulation problem gets considerably more complicated when the tool is 
tethered (possesses a cable). Cables are soft and dynamic obstacles for motion planning -- Their position and 
orientation are considerably hard to compute due to their nature. The cable shape changes according to the robot 
actions, the tension applied to it, and the cable elasticity. These properties cause uncertainty and complicate 
the avoidance of robot-cable and obstacle-cable collisions. Furthermore, when the robot end-effector rotates 
around the tool cable, it can get snarled around the robot hand, producing undesired entanglements and cause 
damage to the robot and the tool. Thus, preventing robot-cable entanglements is an important goal for tethered 
tool manipulation. 

\begin{figure}[!htpb]
  \begin{center}
  \includegraphics[width=.95\linewidth]{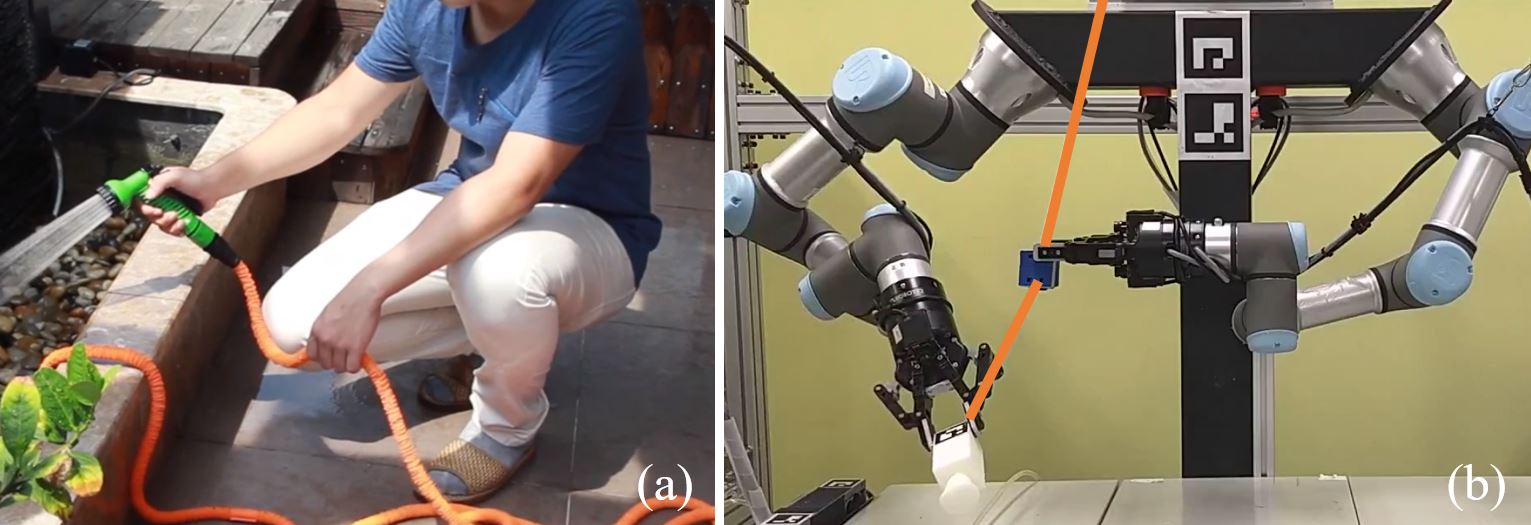}
  \caption{(a) A human handles a tethered tool using both arms: One arm is used for manipulating the tool;
  The other is used for manipulating the cables. (b) The proposed planner motivated by the human strategy.}
  \label{fig:teaser}
  \end{center}
\end{figure}

In this paper, we present a planner for manipulating tethered tools using dual-armed robots. The planner is 
motivated by human manipulation strategies widely seen in our daily life. Fig.\ref{fig:teaser} shows an example. 
The human in the figure handles a tethered tool using both arms: One arm is used for manipulating the tool; 
The other is used for manipulating the cables. Following this strategy, we develop a dual-arm tethered tool 
manipulation planner which generates an Object Manipulation Motion Sequence (OMMS) to handle the tool, and 
a Cable Manipulation Motion Sequence (CMMS) constrain the cable. The planner can prevent entanglement and 
guarantee that the robot (1) avoids bending the tool cable in excess, and (2) avoids cable-robot and 
cable-environment collisions when possible.

Especially, our implementation uses a tool balancer and a cable slider. The balancer simplifies the cable 
deformation problem by constantly applying a pulling force, which forces the cable to form a straight line, 
facilitating the obstacle-avoidance computations. The cable slider is a self-designed mechanical device 
attached to the tool cable. The slider stuck to the cable when it is not held by the robot. When it is 
pressed the slider is released, and it allows the cable to slither through a central hole freely.
Fig.\ref{fig:teaser}(b) shows the robot, its working environment as well as the cable slider (the blue box 
held by the right arm). The tool balancer is not shown, but it is hanged overhead straightening the cable. 
Besides the developed planner, this paper also provides a metric to evaluate the cable state or snarling 
around the robot end-effector, which can be used to compare our solution to other methods. The planner 
guarantees minimal robot-cable contact and avoids undesired entanglements by preventing excess bending. 
Simulations and real-world experiments help validate the presented solution.

\section{Related work and Contribution}

This paper develops a motion planning solution for tethered tool manipulation using dual-armed robots. 
It emphasizes the prevention of entanglements to increase robot safety and manipulation success rates. Accordingly, 
this section reviews related publications on motion planning and manipulation planning, with particular attention 
given to cable-like objects. 

\subsection{Motion planning}

A considerable amount of publications are aimed to develop robot motion planning\cite{masehian2007classic}. 
Early and influential work on motion planning include algorithms for path planning \cite{lozano1979algorithm} 
as well as approaches based on fuzzy logic \cite{vachtsevanos1986fuzzy,walker2000fuzzy}, genetic algorithms \cite{parker1989inverse,gen1997genetic}, 
and neural networks \cite{zacksenhouse1988neural,kozakiewicz1991neural}.

Nowadays, more refined methods for motion planning have been proposed. For example, in \cite{finn2017deep} a model predictive 
control algorithm based on probabilistic inference through a learned predictive image model is presented. The algorithm is 
used to plan for actions that move user-specified objects in the environment to user-defined locations. \cite{ichter2018learning} 
presents a methodology for nonuniform sampling to accelerate sampling-based motion planning. In \cite{solovey2016finding}, 
a discrete RRT algorithm for path planning is shown. \cite{pellegrinelli2017motion} shows an integrated motion planning 
and scheduling method for human and industrial-robot collaboration. 

\subsection{Manipulation planning}

Manipulation planning can be considered as a constrained case of motion planning. The motion sequences generated by 
manipulation planning allow the robot to move objects and to modify its environment's structure \cite{alami1994two}. 
The planning process takes into account the movement of the robot in an environment with movable 
(manipulated) objects in addition to the environment static obstacles.

Recently, manipulation planning has been the focus of several work such as algorithms for 
single-arm and dual-arm object pick-and-place using regrasps \cite{wan2016developing}, a 
probabilistically complete planner for prehensile and non-prehensile actions in cluttered environments \cite{garrett2015backward}, 
an algorithm to preserve object stability under changing external forces \cite{chen2018manipulation}, planning solutions for 
manipulating an elastic object from an initial to a final configuration \cite{lamiraux2001planning}, a planning framework that 
uses non-prehensile actions for the rearrangement of clutter and manipulation of object pose uncertainty \cite{dogar2012planning}, 
and a manipulation planner for the cleaning of planar surfaces \cite{martinez2015planning}.

\subsection{The manipulation planning of cable-like objects}

In particular, motion planning and manipulation planning for handling cables or cable-like objects represents a challenging task. 
Several strategies have been proposed to solve the task. For example, a study on quasi-static manipulation of a planar kinematic 
chain is presented in \cite{mccarthy2012mechanics}. A control solution, for the manipulation of a fire hose, was shown 
in \cite{ramirez2016motion}. A planner for manipulation of interlinked deformable linear objects for aircraft assembly 
was shown in \cite{shah2018planning}. A planning method for knotting/unknotting of deformable linear objects \cite{wakamatsu2006knotting}, 
and a motion planner to manipulate deformable linear objects is described in \cite{saha2008motion}.

\subsection{Contributions}

The work mentioned above presents solutions for manipulating cable-like objects, but they do not address robot-cable entanglement 
avoidance or excessive bending. The definition of cable entanglement can be subjective. In theory, if the robot avoids collision 
with the tool cable, there will be no entanglements. However, in practice, contact between the robot arm and the cable is often 
unavoidable. In such cases, it is important to establish a criterion to differentiate between dangerous cable collisions and 
unavoidable but manageable robot-cable contact. We do so by defining the angle accumulation concept and design a planner 
that tries to diminish the accumulated angle. 

In one of our previous work \cite{sanchez2019arm}, we presented a planning solution for regrasp manipulation of tethered 
tools with tool balancers, but that solution was based on avoiding robot poses or motions that could cause entanglements. 
The solution helped to avoid collisions with the cable but significantly diminished the freedom of movement of the robot. 
Unlike the previous solution, in this work, we generate a motion sequence to manipulate both the cable and the tool and 
diminish cable collisions. The cable maneuvering motions are performed to control the cable bending angle and keep it 
away from others.

\section{Manipulation Planning for Tethered Tools}

\begin{wrapfigure}{r}{0.47\linewidth}
  \centering
  \includegraphics[width=.95\linewidth]{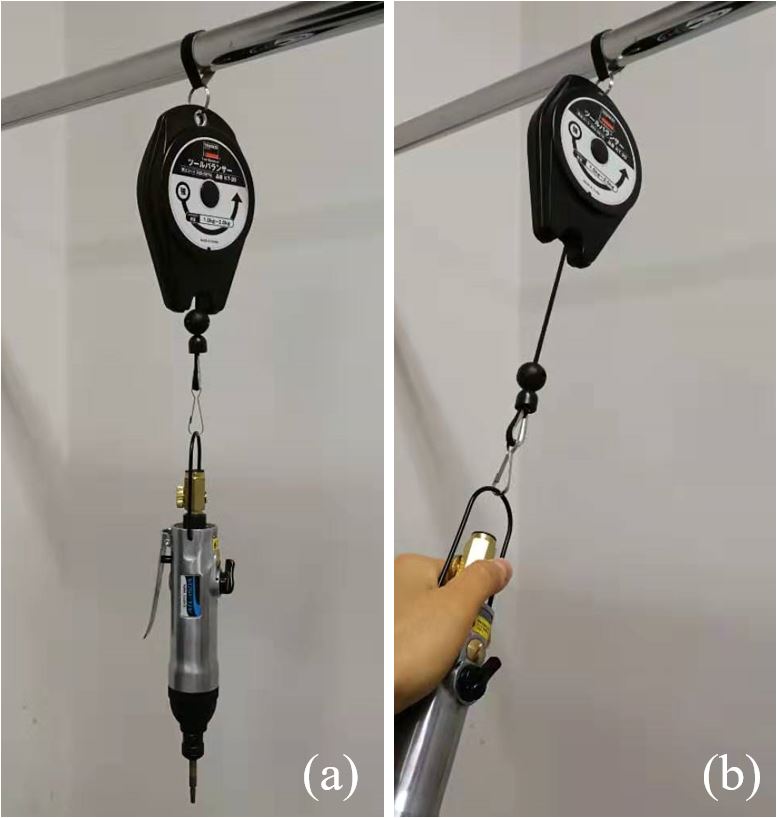}
  \caption{The cable of the tool balancer presents a constant pulling force that simplifies the cable deformation
  into a stright line.}
  \label{fig:toolbalancer}
\end{wrapfigure}

The present method for tool manipulation employs a tool balancer to suspend the manipulated tools. A tool balancer is a 
device that provides a cable to hang tools. The cable presents a constant pulling force that simplifies the cable deformation 
problem by making the cable form a straight line between its endpoint and the tools connection point,
as seen in Fig.\ref{fig:toolbalancer}. 

In our case, the installation of the tool balancer plays a key role in the success of tool manipulation. The initial position 
of the tool balancer was chosen using a manipulability-reachability based rating method. The method is used to determine the 
best starting positions for the balancer. 

The planner computes two motion sequences to realize tethered tool manipulation. The first motion sequence is an Object 
Manipulation Motion Sequence (OMMS), which is computed using our previously proposed single-arm manipulation planner \cite{wan2016developing} 
to manipulate the tool and place it in the desired pose. The second sequence, the Cable Manipulation Motion Sequence (CMMS), 
is used to modify the tool cable shape: The robot manipulates a cable slider to control the bending and the position of the cable. 
The CMMS diminishes the occurrence of robot-cable collisions by placing the cable directly behind the tool during its manipulation. 

\subsection{Cable angle accumulation and entanglement avoidance}



In principle, if the robot avoids any contact between the cable and itself or the environment, 
then it can avoid entanglements entirely. In practice though, these conditions are not always possible 
since the robot end-effector is usually close to the tool cable. Furthermore, slight contact between 
the end-effector and the tool cable can be tolerated if it does not represent a risk for either.

\begin{wrapfigure}{r}{0.47\linewidth}
  \centering
  \includegraphics[width=.95\linewidth]{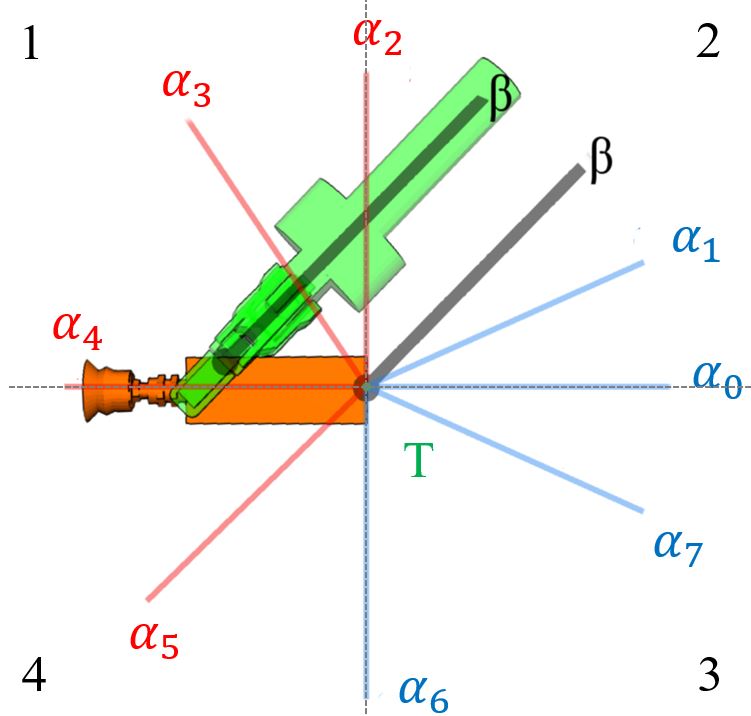}
  \caption{A 2D illustration of angle accumulation. 
  If the cable bending angle (blue and red segments) 
  surpasses the angle of grasping or 90 degrees, the cable starts to get 
  snarled around the end-effector (red segments).
  The arabic numbers indicate the quadrants in the tool reference frame.}
  \label{fig:angleacc}
\end{wrapfigure}

To tackle this problem, we introduce the concept of angle accumulation. Angle accumulation represents 
the extra rotation of the cable around a collision point in the end-effector -- If the tool cable collides 
with the robot end-effector, subsequent rotations around the collision point will cause the cable to get more 
snarled around the robot. Thus, it is important to prevent or diminish angle accumulation. 
Fig.\ref{fig:angleacc} helps illustrate the concept. The figure shows a 2D simplification 
the angle accumulation problem. By representing the tool cable as a straight line (thanks to the tool balancer)
and measuring the angle of bending at the tool reference frame, we can determine if the cable stays within a permissible zone.
If the tool cable bending (measured as 0 when the cable is in the $\alpha_{r}$ reference state, equivalent to the $\alpha_{0}$ 
state in Fig. \ref{fig:angleacc}), stays within the permissible states (its bending angle does not surpass the angle of grasping $\beta$ or 90 degrees), 
we can assume the cable will not collide with the end-effector or the tool itself. If on the other hand, the cable bending goes
beyond the maximum angle $\beta$, we say it is in a state of angle accumulation around the end-effector.
If the cable surpasses the 90 degrees of bending while rotating clockwise (using Fig.\ref{fig:angleacc} as reference)
we say the cable is in a state of angle accumulation around the tool. To calculate the angle accumulation around
the end-effector for a cable state $\alpha_{i}$, we use Eq.\eqref{eq:angleacc} to compute the accumulation magnitude:
\begin{equation}\label{eq:angleacc}
  A_{cc}(i) =   \left \{ \begin{matrix} 0 \mbox{    if $\angle\alpha_{r}\textit{T}\alpha_{i} < \beta$, else }& \mbox{ }\mbox{ \mbox{}}
\\ A_{cc}(i - 1) + \eta_{\phi}(\alpha_{i})\{\angle\alpha_{r}\textit{T}\alpha_{i} - \angle\alpha_{r}\textit{T}\alpha_{i-1}\}-\beta  & \mbox{}\end{matrix}\right.  
\end{equation}

Here, $\alpha_{r}$ represents the cable with no bending, equivalent to $\alpha_{0}$ in Fig.\ref{fig:angleacc}. 
$\textit{T}$ is the connection point between the cable and the tool. $\alpha_{i}$ is the $i$-th cable state, 
which is represented as a vector or a straight line that goes from the tool tail to the cable anchor point 
(tool balancer or cable slider). Basically, the function adds up the differential changes 
in angle between cable states $\alpha_{i}$ and the reference state $\alpha_{r}$. The value of $\eta_{\phi}(\alpha_{i} ))$ 
acts as a memory variable with values that depend on the cable position, based on the quadrants at the tool local reference frame, 
the value of $\eta_{\phi}(  \alpha_{i})$ is described by Eq.\eqref{eq:etavalue}:
\begin{equation}\label{eq:etavalue}
\eta_{\phi}(\alpha_{i}) = \left \{ \begin{matrix} 1 & \mbox{if }\mbox{  $\alpha_{i}\in$\mbox{ quadrants 1 or 2}}
\\ -1 & \mbox{if }\mbox{  $\alpha_{i}\in$\mbox{ quadrants 3 or 4}}\end{matrix}\right.  
\end{equation} 

The variable $\eta_{\phi}(\alpha_{i})$ is used to correct the addition of differential accumulation 
when the cable enters quadrants 3 and 4 shown in Fig.\ref{fig:angleacc}. For example, If the cable starts at 
state $\alpha_{0}$ and rotates counter-clock wise, we must adjust the sign of the magnitude 
$\angle\alpha_{0}\textit{T}\alpha_{i} - \angle\alpha_{0}\textit{T}\alpha_{i-1} $ after the cable surpasses state $\alpha_{4}$, 
otherwise, the differential sum $\angle\alpha_{0}\textit{T}\alpha_{i} - \angle\alpha_{0}\textit{T}\alpha_{i-1} $ will 
be negative and the angular accumulation will decrease when, in reality, the cable is getting more snarled around the end-effector. 

To calculate the angle accumulation around the tool, (for clock-wise rotations, using Fig.\ref{fig:angleacc} 
as reference) we can use:
\begin{equation}\label{eq:angleacc2}
  A_{cc}(i) =   \left \{ \begin{matrix} 0 \mbox{    if $\angle\alpha_{r}\textit{T}\alpha_{i} < 90$, else }& \mbox{ }\mbox{ \mbox{}}
\\ A_{cc}(i - 1) + \eta_{\theta}(\alpha_{i})\{\angle\alpha_{r}\textit{T}\alpha_{i} - \angle\alpha_{r}\textit{T}\alpha_{i-1}\}-90  & \mbox{}\end{matrix}\right.  
\end{equation} 
where $\eta_{\theta}=-\eta_{\phi}$. To determine if the angle accumulation is around the tool or the end-effector, 
we can check in which quadrant the cable is located when its bending surpasses 90 degrees or $\beta$ respectively.

\subsection{High manipulability region}
The initial installation position of the tool balancer is an important factor in successful manipulation planning. 
A non-optimal installation position can pose the tool and cable slider in difficult to grasp positions,
hindering the planner ability to find a solution for a given manipulation task.  

Intuitively, the effectiveness of our planner depends on the initial position of the tool balancer. 
If the balancer is too far from the robot, the number of IK-feasible grasps of the cable slider and the tool decreases. 
On the other hand, if the tool balancer is too close to the robot, the maneuvering space for the tool cable could be 
too small, increasing the risk of robot self-collision. Also, the robot should be able to grasp the tool and the cable 
using both arms, so the balancer must be centered in front of the robot.

Following these considerations, we optimize the initial installation position of the tool balancer by using a 
rating method based on the robots reachability and manipulability. Our rating method computes the most advantageous 
positions for the tool balancer and is also used to compute a region of high manipulability. 

\subsubsection{Grasp-based reachability region for dual armed robots}

An optimal tool balancer position can help the planner by (1) placing the tool and its cable in a position of 
high reachability and manipulability, and (2) giving the robot enough room to maneuver the tool and the cable. 
We employ a reachability test for the robot workspace to find balancer positions that comply with both requirements, 
to find an optimal balancer position. 

Firstly, we map the workspace into several points in a grid separated by 50 $mm$ each. Secondly, we tasked our 
IK-solver to generate IK solutions to place the robot end-effectors in every point of the grid. The points 
with at least one IK solution are cataloged as reachable.

These reachable points are then used to map the robot workspace with regions of reachability $\Omega_{r}$ and $\Omega_{l}$ 
for the right and left arms respectively. A region of dual-arm reachability can be computed by intersecting $\Omega_{r}$ 
and $\Omega_{l}$ like in Eq.\eqref{eq:reachabilityregion}:
\begin{equation}\label{eq:reachabilityregion}
\Omega = \Omega_{r}\cap\Omega_{l}
\end{equation}

The resulting region, $\Omega$ contains candidate positions for the tool placement. Since the tool 
balancer is used to hang the tool vertically, once the horizontal or $\textit{\textbf{x}}$ and $\textit{\textbf{y}}$ 
coordinates of the balancer (in the robot reference frame) are set, the tool and the cable slider can only change their
resting position in the vertical axis. We aim to find a position in the robot horizontal plane 
($\textit{\textbf{x}}$-$\textit{\textbf{y}}$ plane) that maximizes the dual-arm reachability in the robot vertical axis.

To choose the optimal $\textit{\textbf{x}}$ and $\textit{\textbf{y}}$ coordinates for the balancer position, 
we evaluate each $\textit{\textbf{x}}$-$\textit{\textbf{y}}$ coordinate pair in our grid by counting the 
reachable points in their vertical axis. That is, we fix the $\textit{\textbf{x}}$ and $\textit{\textbf{y}}$
coordinates and test the reachability of the points in the vertical axis $\textit{\textbf{z}}$, increasing the
height 50 $mm$ at a time. The $\textit{\textbf{x}}$-axis in the robot reference frame points to the robot front,
the $\textit{\textbf{y}}$-axis points to the robot left-hand side and the $\textit{\textbf{z}}$ -axis points upwards. 

From our analysis, the coordinate pair $(300,0)[mm]$ (300 $mm$ in front of the robot, centered between its arms) yielded
the highest amount of evaluated reachable points along the vertical axis $\textit{\textbf{z}}$, with 15 reachable points.
Nonetheless the $\textit{\textbf{x}}$-$\textit{\textbf{y}}$ coordinate pair is dangerously close to the robot frame,
which significantly impairs our CMMS since the CMMS normally places the cable-holding arm behind the tool-handling arm, 
leading to very close motion and a small zone near the robot body. To avoid these problems, we decided to place the 
tool balancer farther in front of the robot, at the coordinates $(450, 0, 1800)[mm]$ at the cost of selecting a 
pair of $\textit{\textbf{x}}$-$\textit{\textbf{y}}$ coordinates with 14 reachable points along the $\textit{\textbf{z}}$-axis.

\subsubsection{Grasp-based manipulability analysis}

The CMMS involves motions to make the cable-holding arm follow the movement of the tool while grasping a cable slider. 
Usually, these motions place the cable-holding arm in a region between the robot-body and the tool-handling arm. 
In this region the arm robot requires high mobility to avoid collisions and complete its task. 
To ensure high mobility for the cable-holding arm, we would like to place the cable slider 
in a region with a high expected manipulability index.

The manipulability of a robot represents its ability to position and re-orientate its end-effector given an 
initial joint configuration. While inspecting the infinite joint configurations of the robot arm is impractical, 
we can perform a grasp-based analysis to obtain an expected or average manipulability based on the cable slider 
position and the IK-feasible grasps for that particular position. The result is an average value of 
manipulability associated with a point in space.

To compute the expected manipulability of a point in space, we place the slider in this point using our simulation 
environment. Then, we compute the IK solutions if they exist, that allows the robot to grasp the tool using 
the grasps stored in our grasp database \cite{wan2017iros}. For simplicity, the slider orientation is fixed to a single 
value when evaluating the possible grasps. Our algorithm then evaluates the manipulability of each Ik solution and 
computes the median value for a single point in the robot workspace using Eq.\eqref{eq:graspmanipulability}:
\begin{equation}\label{eq:graspmanipulability}
  M(p(x,y,z)) = \frac{\sum\limits^{G(p(x,y,z))}_{n=0}{m(g_{n}(p(x,y,z)))}}{G(p(x,y,z))}
\end{equation}

\begin{figure}[!htpb]
  \begin{center}
  \includegraphics[width=.95\linewidth]{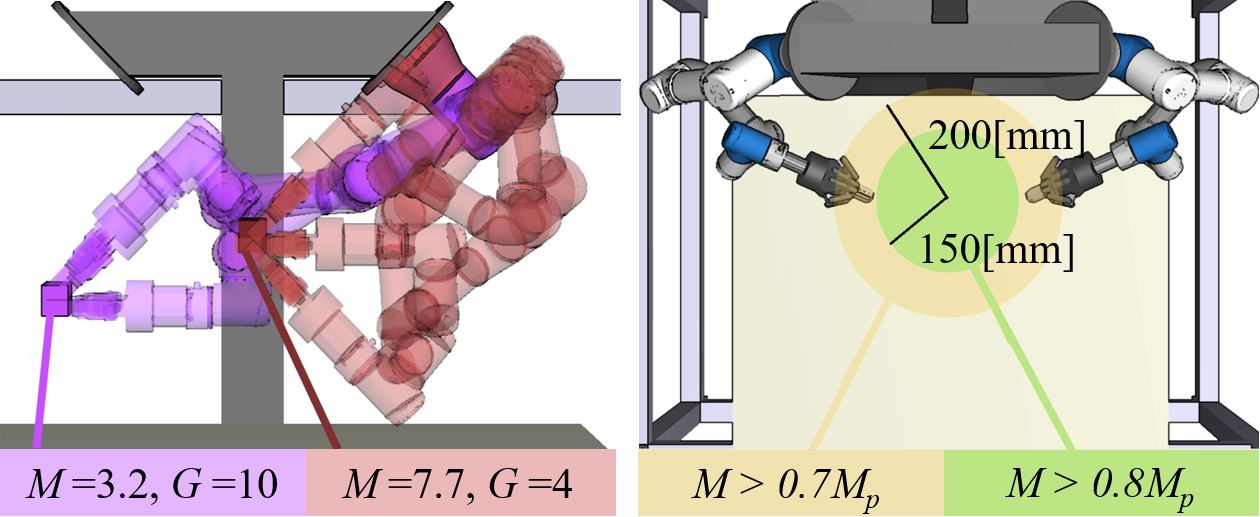}
  \caption{Manipulability Analysis. Left: We compute the different grasps of the tool balancer for two cable slider positions. 
  The manipulability for each grasp is calculated and added to a total value. The position with the highest average 
  manipulability $M$ and possible grasps $G$ is a more desirable starting position for the slider. 
  Right: Manipulability spheres. Within the radius of the spheres, the values M and G stay above threshold values 
  based on a position of high manipulability and high amount of IK-feasible grasps.}
  \label{fig:Manipulability Analysis}
  \end{center}
\end{figure}

Here, $M$ is the average manipulability score for a point $p(x,y,z)$, $G(p(x,y,z))$ is the total (non-zero) amount 
of IK-feasible grasps for the slider in point $p(x,y,z)$, $m$ is a function that returns the manipulability of the 
robot based on its joint angles and its maximum angles of rotation and $g_{n}$ is the $n$-th set of joint angles 
that place the robot end-effector in the pose necessary to execute the $n$-th grasp. Fig.\ref{fig:Manipulability Analysis} 
shows the process of calculating the manipulability for two different slider positions.

Using Eq.\eqref{eq:graspmanipulability}, we computed the average manipulability and available grasps within the grid 
used in the previous sub-section. The highest average manipulability scores were registered within a certain region of
the robot workspace. By using these scores, we can create a ``manipulability sphere'' to represent this region in which the number of 
possible slider grasps $G(p(x,y,z))$ and the score $M$ stay above certain reference values. We used the coordinates $(400,0,1450)[mm]$ as 
our reference point since it has a central location in the robot reachable zone, the coordinates yielded a $M_{p}$ value 
for the average manipulability and $G_{p}$ number of unique Ik-feasible grasps. 

Afterward, we explored the remaining points in the grid and realized that, by keeping the slider within a 150 $mm$ 
radius from the reference point, the manipulability and available grasps of the evaluated points stay above $0.8M_{p}$ and $0.5G_{p}$. 
Since the chosen tool balancer position will directly place the slider at the coordinates $(450, 0, z)[mm]$, where the 
height $z$ is variable, we can assume the initial position where the slider will most likely be within $150[mm]$ of the 
reference point and a relatively high manipulability for the initial grasp can be expected. Fig.\ref{fig:Manipulability Analysis} 
shows a representation of the manipulability sphere of $150[mm]$ and $200[mm]$ and the minimum $M$ and $G$ values registered within these regions.

\subsection{Object manipulation planning}

The OMMS is generated in three steps using our previous planner \cite{wan2016developing}. Firstly, 
the planner selects a candidate object grasp ${}^{}_{}{C}^{\text{h}}_{}$ from a previously-built database to 
pick-up the object in a starting pose. The grasp database \cite{wan2017iros} is computed offline in the object's 
local coordinate system $\Sigma_{\text{t}}$. Secondly, the planner checks the IK-feasibility and robot-object 
collisions of the start and goal robot grasping poses. The grasping poses are represented by a given end-effector 
transformation matrix ${}^{\text{o}}_{}{T}^{}_{}$ which can be computed using Eq.\eqref{eq:graspingpose}:
\begin{equation}\label{eq:graspingpose}
  {}^{\text{o}}_{}{T}^{}_{} = {}^{}_{}{C}^{\text{h}}_{}   {}^{}_{}{O}^{\text{o}}_{}
\end{equation}

Here, ${}^{}_{}{O}^{\text{o}}_{}$ represents the transformation matrix of the tool for a given pose 
(starting or goal object pose) in the robots reference frame $\Sigma_{\text{o}}$. Finally, the planner 
connects the starting and goal grasping poses of the object through an intermediate/transfer robot poses generated by an
RRT-based sampling method. The result is a series of motions that allow the robot to grasp an object, maneuver it through
its workspace, and place it in the desired goal pose. The object movement associated with this motion sequence is subsequently used to plan the CMMS.

\subsection{Cable manipulation planning}

\begin{figure}[!htbp]
  \begin{center}
  \includegraphics[width=.95\linewidth]{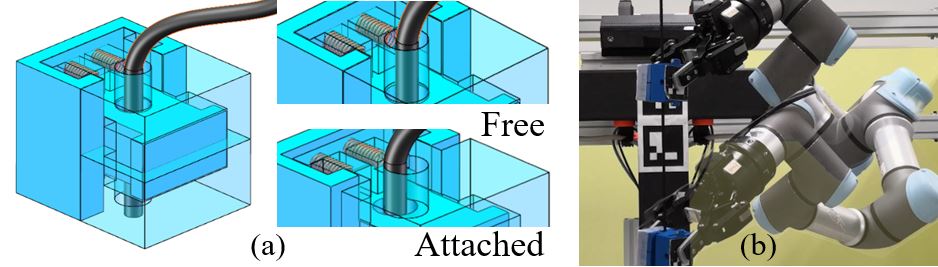}
  \caption{(a) The cable slider model. (a-Free) The slider in its free state. 
  When the robot gripper grasps the slider, the gripper overcomes the spring forces and pushes the two slider 
  internal circular holes to a concentric state, allowing for the free movement of the cable through both holes.
  (a-Attach) The slider in its attached state. When it is not being manipulated, its internal springs apply 
  a constant force, constraining the cable. (b) The robot manipulating the slider in its free state.} 
  \label{fig:sliderhandle4imgf}
  \end{center}
\end{figure}

The CMMS is computed to control the cable-manipulating arm and place the cable in optimal positions that diminish cable angle
accumulation around the end-effector and prevent robot-cable collisions. The robot handles the cable using a slider tool,
as seen in Fig \ref{fig:sliderhandle4imgf}.

The cable can slide through the slider, simplifying the cable manipulation problem to a slider placement problem.
To generate a CMMS and avoid entanglements, our planner selects one of the possible slider grasps and generates
the motions necessary to reach the selected grasp. Subsequently, the planner computes the tool motions associated
with the OMMS and estimates the optimal cable positions for every intermediate state of the tool generated by
RRT-based exploring. The result is a motion sequence that allows the robot to reach and grasp the cable slider 
and control the cable movement, placing it directly behind the tool if possible, preventing collisions and excess
angle accumulation between the end-effector and the cable.

\begin{figure}[!htbp]
  \begin{center}
  \includegraphics[width=.95\linewidth]{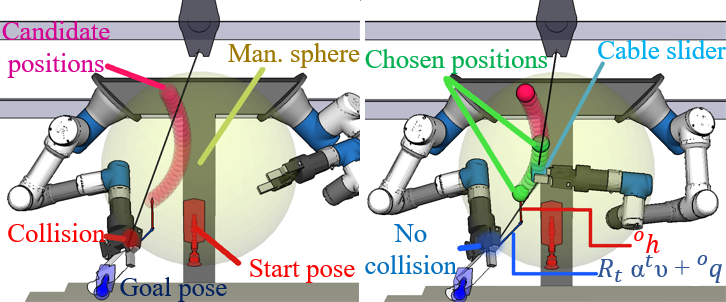}
  \caption{Simulated OMMS-only and OMMS+CMMS tool manipulation. Left:
  The robot executes an OMMS with its right arm. The planner uses the vectors 
  ${}^{\text{o}}_{}{R}^{}_{t}\alpha{}^{\text{t}}_{}{\boldsymbol{\mathit{\upsilon}}}^{}_{} + {}^{\text{o}}_{}{q}^{}_{} $ (in blue) 
  and ${}^{\text{o}}_{}{h}^{}_{}$ (red) to calculate candidate positions for the cable slider 
  (represented by the purple spheres) by using the vectors. Right: 
  The robot moves the cable slider to selected candidate positions. 
  The chosen positions (green spheres) minimize the bending angle of the cable and prevent collisions.
  Abbreviation: Man. sphere -- Manipulaiton sphere.} 
  \label{fig:simulationsetting}
  \end{center}
 \end{figure}

The OMMS is used to calculate the poses of the tool during the manipulation process and generate the CMMS. For each tool pose,
the planner computes a projection from the tool's tail (connection point between the tool and its cable). The projection will
be used as goal positions for the slider tool. Each projection position $ {}^{\text{o}}_{}{p}^{}_{}$ (as described in
the robots reference frame $\Sigma_{\text{o}}$) can be computed using 
Eq.\eqref{eq:cablepositionvec}:
\begin{equation}\label{eq:cablepositionvec}
{}^{\text{o}}_{}{p}^{}_{} = {}^{\text{o}}_{}{R}^{}_{t}\alpha_{s}{}^{\text{t}}_{}{\boldsymbol{\mathit{\upsilon}}}^{}_{} + {}^{\text{o}}_{}{q}^{}_{} + {}^{\text{o}}_{}{{h}}^{}_{},
\end{equation}
where ${}^{\text{o}}_{}{R}^{}_{t}$ represents the tools rotation matrix. ${}^{\text{t}}_{}{\boldsymbol{\mathit{\upsilon}}}^{}_{}$ 
is an unitary vector in the tool's reference frame $\Sigma_{\text{t}}$. It points to the tool tail normal direction. 
The scalar value $\alpha_{s}$ dictates the magnitude of the projection or how far behind the tool the cable slider 
should be placed. The object's position ${}^{\text{o}}_{}{q}^{}_{}$ is added to place the projection in the correct 
position in the robot's reference frame. Finally, the vector ${}^{\text{o}}_{}{h}^{}_{}$ is added to translate 
the projection point vertically in order to maintain a minimum height for the slider position (to avoid collisions 
with the table). Fig.\ref{fig:simulationsetting} better illustrates this process. The planner then examines every 
point $ {}^{\text{o}}_{}{p}^{}_{}$ as a candidate goal position for the cable-holding arm. 
 
\textit{Discarding slider positions:}
Each slider position is linked to its corresponding object pose and the pose of the tool manipulating arm 
(dictated by the OMMS), to form a list $P$ of candidate slider poses. The corresponding robot poses, 
object position, and slider candidate position are used to verify collision avoidance. To perform 
collision detection, we assume the cable shape is represented by two straight lines, the first line 
goes between the tool and the slider, and the second, between the slider and the balancer, both lines 
can be represented as vectors. With the line vectors, we can check if the robot collides with the cable 
during manipulation. Also, we can measure the angle between the cable (the section that goes from the tool to the slider) 
and the end-effector to verify if there is angle accumulation.   

Ideally, the robot can place the slider in all its candidate position points without collisions. If there are points 
in $P$ that are either, not reachable by the robot, cause cable collisions (disregarding the robot end-effector), 
or surpass the angle accumulation threshold, the planner discards them and uses RRT exploring to connect the closest 
adjacent points that do not violate these conditions. 

In the case the planner does not find a motion sequence that preserves the angle accumulation below 
a given threshold (30 degrees in our case) for all the robot states, the planner can compute the goal 
positions again by reducing $\alpha_{s}$ by 20\%. If it fails to find a solution again, the planning fails and 
a new OMMS must be computed again to find an alternative CMMS.

\section{Experiments and Analysis}\label{sec:experiment}
\subsection{Angle accumulation measurements}

Several benchmarks to test the performance of the proposed planner using our simulation environment\footnote[1]{For more information
about our planner and simulation environment, please visit: https://gitlab.com/wanweiwei07/wrs\_nedo } are performed. 
For these tests, the robot right arm manipulates the object, and the left-hand maneuvers the cable, the threshold for
maximum angle accumulation is set to 30$^{\circ}$. 

\begin{figure}[!htpb]
  \begin{center}
  \includegraphics[width=.95\linewidth]{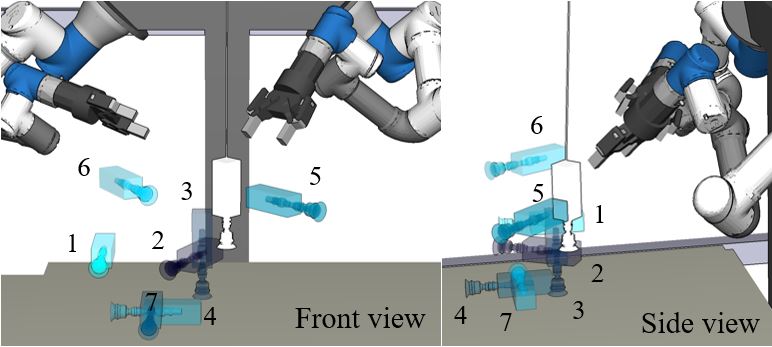}
  \caption{Goals for the simulation benchmarks.
  The blue transparent objects represent the candidate goal poses of the tool. 
  For each benchmark, the robot must grasp the object in its initial pose (white), 
  and place it in three of the randomly chosen goal poses.} 
  \label{fig:7goals}
  \end{center}
\end{figure}

Each benchmark consists of an initial tool pose and three-goal poses. The planner generates motion sequences
to pick up the tool and then place it in the desired goal poses. Seven different goal poses are considered
to create a benchmark. The poses are shown in Fig.\ref{fig:7goals}. For each simulation, we track the angle
accumulation of the robot right arm for later analysis. Also, we used the OMMS-only planner and a planner 
that uses object handover to complete the same tasks and compare the planners. 
 
Table \ref{Tab: Benchmarks} shows the goals chosen for each benchmark (benchmark number in bold and in parenthesis) 
and the maximum and mean angle accumulation (mean in bold and parenthesis) for each solution.

\begin{table}[!htbp]
  \begin{center}
    \setlength{\tabcolsep}{4.3pt}
    \caption{\label{Tab: Benchmarks}}
    \begin{threeparttable}
      \begin{tabular}{cc}
        \toprule
        Goals &$A_{cc}$ \textbf{[degrees]}
        \\
        \midrule
        \midrule
        \textbf{(1)} 1, 6, 3 &O+C = 27.61 \textbf{(6.06)}, O = 42.43 \textbf{(16.88)}, H = 83.75 \textbf{(27.14)}\\
        \midrule
        \textbf{(2)} 2, 1, 3 &O+C = 16.56 \textbf{(2.89)}, O = 44.59 \textbf{(17.88)}, H = 44.59 \textbf{(14.21)}\\
        \midrule
        \textbf{(3)} 3, 4, 5 &O+C = 16.49 \textbf{(2.20)}, O = 45.0 \textbf{(22.16)}, H = 82.48 \textbf{(19.95)}\\
        \midrule
        \textbf{(4)} 4, 1, 5 &O+C = 16.70 \textbf{(2.31)}, O = 44.96 \textbf{(24.38)}, H = 80.65 \textbf{(23.55)}\\
        \midrule
        \textbf{(5)} 7, 6, 2 &O+C = 29.83 \textbf{(6.02)}, O = 44.12 \textbf{(23.64)}, H = 75.12 \textbf{(26.95)}\\
        \midrule
        \textbf{(6)} 5, 4, 1 &O+C = 25.05 \textbf{(4.48)}, O = 43.89 \textbf{(27.25)}, H = 44.99 \textbf{(18.74)}\\
        \midrule
        \bottomrule
      \end{tabular}
      \begin{tablenotes}
      Meanings of abbreviation: O+C represents the OMMS+CMMS solution for the given benchmark. O is the OMMS-only solution.
      H represents the solutions using handover.
      \end{tablenotes}
    \end{threeparttable}
\end{center}
\end{table}

The CMMS allows the robot to handle the tool cable to follow the movement of the tool, reducing cable bending 
at the tool local reference frame. On average, the OMMS+CMMS executions reduced maximum angle accumulation by 
50\% when compared to the OMMS-only planner and by 67\% when compared to dual handed tool manipulation. 
The average angle accumulation for the OMMS+CMMS manipulation tasks was also considerably lower for the OMMS+CMMS 
planner, which hints at a lower chance of end-effector-cable entanglements.

\subsection{OMMS+CMMS without the tool balancer}
A downside of the tool balancer is that it does not allow for the regrasping of the tool using
table placements (the cable would pull the tool out of position). If regrasping is necessary, 
the robot is forced to use handover motions, the motions with the highest angle accumulation. 
The proposed planner can also be used to maneuver tethered tools without the use of a tool balancer,
allowing the tool handling arm to perform regrasps using table placements. 

In this case, one end of the cable is fixed to a corner of the robot table to approximate 
the cable shape as two straight lines, which go between the tool and the cable slider and between 
the cable slider and the fixed point. 

By manipulating the tool cable the robot can not only diminish angle accumulation and the possibility 
of entanglements, as shown in the previous experiment, but it can also maneuver the cable above obstacles in the robot workspace. 

Obstacle-avoidance experiments are performed to test the balancer-less planner. The experiments consist of randomly placing
a box as an obstacle in the robot workspace and performing a manipulation task with a tool starting pose, 
and two-goal poses. The planner is tasked to place the tool in two-goal positions, and the amount of 
cable-obstacle collisions are measured for each planner. Ten different tests are performed with random 
box positions using our planner and the OMMS-only planner. The straight-line cable approximation is used
to detect collisions in our simulation environment. Real tests are performed to assess collisions with the obstacles.

\subsection{Real-world experiments}

After testing our planner in simulations, we applied our solution to our real-world robot. 
The robot uses its hand-mounted cameras to detect the AR Markers on the tool and the slider and 
compute their current pose. For these experiments, we tested the same motion sequences planned 
in our simulations. 

\begin{figure*}[!htpb]
  \begin{center}
  \includegraphics[width=.95\textwidth]{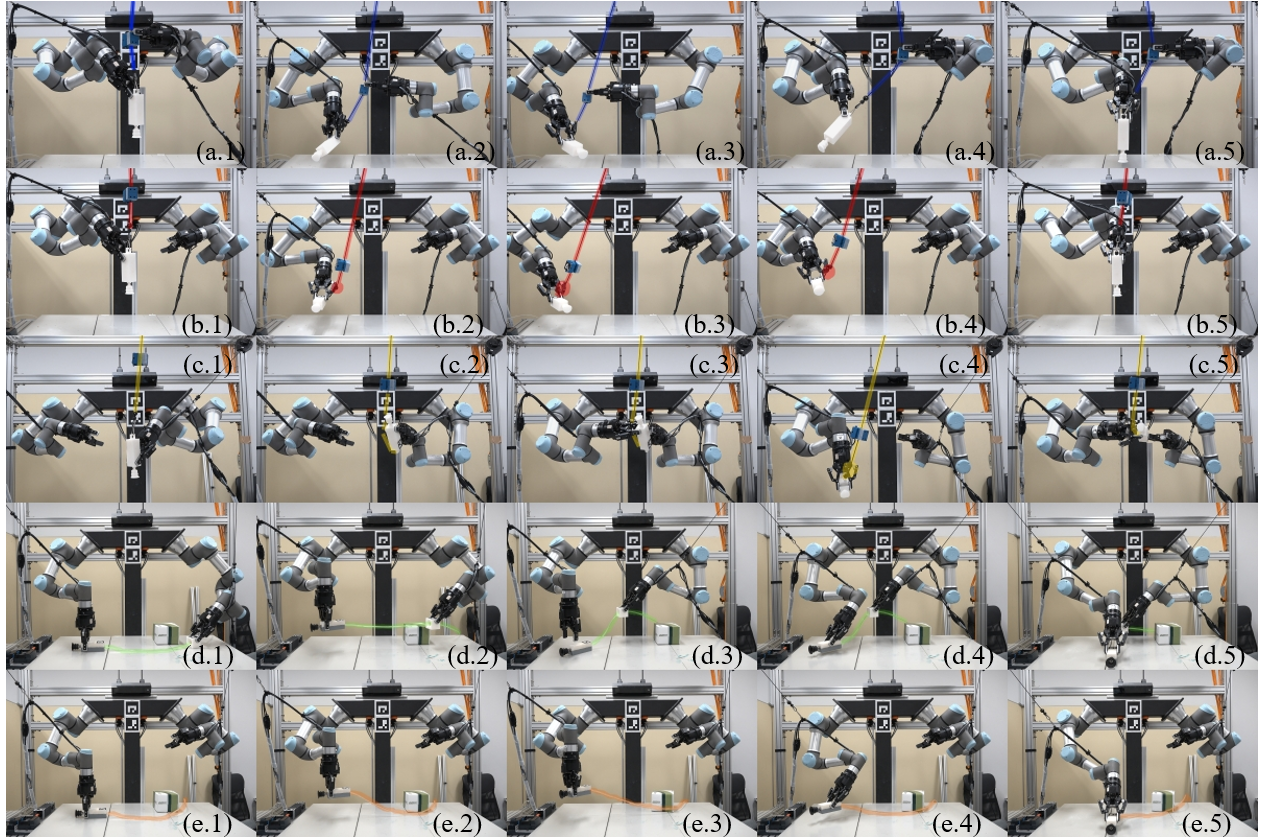}
  \caption{Real-world implementations. The first row shows the OMMS+CMMS motion sequences --
  The robot completes its task without surpassing the maximum angle accumulation threshold of 30$^{\circ}$. 
  The second row shows the OMMS-only sequence, from the second to the fourth image we can observe the excess bending on the cable. 
  The third row shows a part of the planner solution involving handover.
  In this case, the cable gets snarled around the robot end-effector when the robot performs the handover motion. 
  Rows four and five show the OMMS+CMMS and OMMS-only solutions respectively. 
  In the OMMS+CMMS solution, the robot successfully maneuvers the cable and avoids the obstacle by lifting it above the box. 
  The OMMS-only solution, on the other hand, does not consider the box or the cable and results in a cable-box collision. } 
  \label{fig:Real Exps}
  \end{center}
\end{figure*}

In all cases, the robot was able to complete the OMMS+CMMS task while manipulating the cable. 
In Fig.\ref{fig:Real Exps} a real-world execution of the planner performing benchmark 1 can be seen 
and compared to the regular OMMS-only planner and the solution provided by the planner using handover. 
The angle accumulation comparison between planners for benchmarks 1 through 6 is shown in Fig.\ref{fig:123456 exps}. 
A video demonstration can be seen in the supplementary material.

Furthermore, we also executed the cable-box collision avoidance motion sequences. An example can also be seen in 
Fig.\ref{fig:Real Exps}. Table \ref{Tab: Obstacle Avoidance Test} shows the results of real-world executions.
The presented planner avoids collisions in 70\% of the cases while the OMMS-only planner is only successful in 20\% of the cases. 
The cable shape is assumed to be a straight line for the OMMS + CMMS planner since the robot holds the cable slider close to the tool. 
The same approximation cannot be used for the balancer-less, OMMS-only solution since the cable does not form a straight line, making its 
shape difficult to predict. A video demonstration can be seen in the supplementary material. 

\begin{table}[!htbp]
  \begin{center}
  \setlength{\tabcolsep}{7.3pt}
  \caption{\label{Tab: Obstacle Avoidance Test}}
  \begin{tabular}{cccc}
  \toprule
  Planner & Success &  Collisions & Mean $A_{cc}$\textbf{[degrees]}
  \\
  \midrule
  \midrule
    OMMS& 2 & 8 & N/A\\
  \midrule
    OMMS + CMMS & 7 & 3 & 6.81\\
  \midrule
  \bottomrule
  \end{tabular}
  \end{center}
\end{table}

\begin{figure*}[!b]
  \begin{center}
  \includegraphics[width=.95\textwidth]{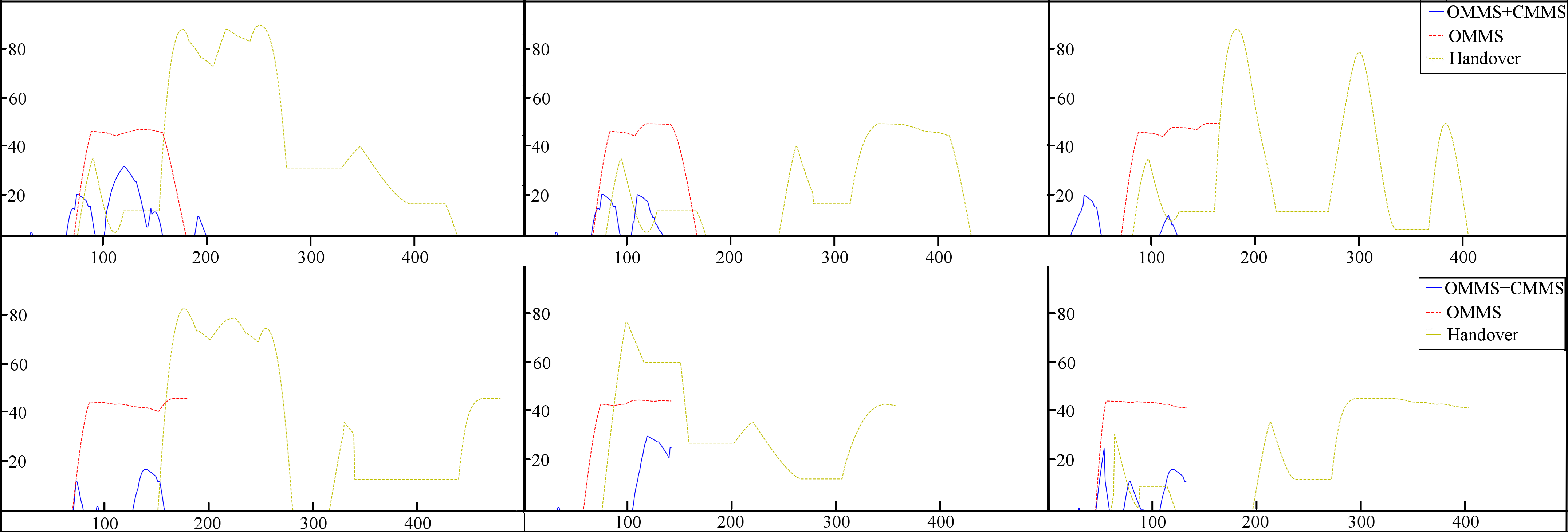}
  \caption{Angle accumulation for the benchmarks, from the top left to bottom 
  right graphs the figure shows the accumulation angles registered for 
  benchmarks 1 through 6 respectively. The vertical axis shows the angle 
  accumulation in degrees. The horizontal axis indicates the robot n-th robot 
  state during the execution. The blue lines show the results for our OMMS+CMMS planner, 
  the red lines show the angle accumulation for the OMMS-only planner and the 
  yellow lines show the accumulation for the solutions with object handover. } 
  \label{fig:123456 exps}
  \end{center}
\end{figure*}

\section{Conclusions}

In this paper, we presented a manipulation planner for entanglement avoidance. Simulations and
real-world experiments confirm that the planner generates motion sequences that reduce angle accumulation 
around the robot end-effector and allow collision avoidance between the cable and the robot and the cable
and its environment. 
The tool balancer provides a constant pulling force for the cable, straightening its shape and simplifying 
collision detection and the computation of the cable shape. On the other hand, the balancer limits the 
regrasping capabilities of the robot.
Furthermore, the experiments without the tool balancer showed that the CMMS allows the robot 
to manipulate the cable to avoid obstacles and also perform tool regrasping by using placements. 
Cable obstacle avoidance can be especially useful in cluttered environments where the cable could 
push objects outside of the robot reach or disturb the original positions of obstacles and objects, 
which can be fatal for offline manipulation planners like ours. Nonetheless, a more accurate representation 
of the cable catenary shape could allow a higher success rate with an increased amount of obstacles. 


Our planner provides a safe alternative to tethered tool manipulation. It reduces cable bending
at the tool local reference frame and angle accumulation during the manipulation task. The use of 
a tool balancer facilitates and makes more accurate the simulations during the planning stage, but
the planner can still be implemented without the balancer to perform tool placements and regrasping. 
Future implementations of the proposed planner will aim to solve the cable deformation problem and
discard the need for a tool balancer.
 
\bibliographystyle{IEEEtran}
\bibliography{paperDaniel}

\end{document}